\documentclass[runningheads]{llncs}

\usepackage{amsmath}
\usepackage{amssymb}
\usepackage{amsfonts}

\usepackage{booktabs}
\usepackage{graphicx}

\usepackage{caption}
\usepackage{subcaption}

\usepackage{listings}
\usepackage[linesnumbered,lined,boxed,commentsnumbered]{algorithm2e}
\usepackage{framed}

\usepackage{xcolor}

\definecolor{applegreen}{rgb}{0.55, 0.71, 0.0}

\usepackage{hyperref}
\hypersetup{
  colorlinks=true,
  citecolor=applegreen
}
\usepackage[nameinlink]{cleveref}

\begin{document}
\title{Random Projections with Optimal Accuracy}
\author{Maciej Skorski}
\authorrunning{M. Skorski}
\institute{University of Luxembourg}
\maketitle              % typeset the header of the contribution
\begin{abstract}
This work constructs Jonson-Lindenstrauss embeddings with best accuracy, as measured by variance, mean-squared error and exponential concentration of the length distortion. Lower bounds for any data and embedding dimensions are determined, and accompanied by matching and efficiently samplable constructions (built on orthogonal matrices).
Novel techniques: a unit sphere parametrization, the use of singular-value latent variables and Schur-convexity are of independent interest. 
%This construction, in particular, achieves best possible exponential (Bernstein-type) confidence bounds.
\end{abstract}
\section{Introduction}
\subsection{Background}

Since the seminal result of Johnson and Lindenstrauss~\cite{johnson1984extensions}, the technique of \emph{random embeddings} has become not only an object of theoretical studies but the fundamental high-dimensional data preprocessing technique~\cite{fradkin2003experiments,bingham2001random,ailon2009fast,ailon2006approximate,lv2008fast}.

In strict statistical terms, the random embedding technique refers to a family of results which show
that random (suitably sampled) linear maps nearly preserve the euclidean distance and, hence, the data structure.
More precisely, the focus of this paper is on \emph{Distributional Johnson-Lindenstrauss} transforms,
suitably sampled $n\times m$ matrices $A$ such that for any $m$-dimensional vector $x$
\begin{align*}
    \Pr_A\left[ 1-\epsilon\leqslant \frac{\|Ax\|_2}{\|x\|_2} \leqslant 1+\epsilon \right] \geqslant 1-\delta.
\end{align*}
Good samplers aim, given fixed data dimension $m$, to maintain low error $\epsilon$ (small distortion), 
low failure probability $\delta$ (high-confidence), while keeping the embedding dimension $n$ much smaller than the data dimension $m$.

In the long-line of research~\cite{frankl1988johnson,indyk1998approximate,achlioptas2003database,ailon2006approximate,matouvsek2008variants,dasgupta2003elementary,dasgupta2010sparse,kane2014sparser,li2006very,freksen2018fully,cohen2016nearly,rojo2010improving} several matrix samplers have been investigated, and significant effort has been made to improve the statistical guarantees. However, given the current state-of-the-art, \emph{statistical accuracy in the finite-sample case is poorly  understood}. The key issues are:
\begin{itemize}
\item \textbf{Asymptotic formulas with big hidden constants.} In nearly all analyses offer only asymptotic formulas, except few research papers~\cite{dasgupta2003elementary,achlioptas2003database,rojo2010improving,burr2018optimal}.
Unfortunately, the dependencies involve large constants~\cite{freksen2018fully,krahmer2011new} and have \emph{exponential impact} on confidence intervals in statistical inference. This applies also to textbooks and expository notes: they offer modular proofs but at the expense of worse bounds~\cite{boucheron2003concentration,dirksen2016dimensionality}. Also impossibility results
are also known only in the asymptotic, rather finite-sample, regime~\cite{burr2018optimal,larsen2016johnson}.
\item \textbf{Theory far behind practice.} Provable formulas are much worse than observed performance, as found in surveys~\cite{venkatasubramanian2011johnson} and when benchmarking~\cite{freksen2018fully}. Theoretical analyses in accompanying experiments often resort to Monte Carlo estimations, finding their formulas inaccurate~\cite{freksen2018fully,jagadeesan2019understanding}.
\item \textbf{Refining fairly difficult}. The prior approaches build less or more directly on the 
quadratic chaos estimation technique, known to be inherently hard and sub-optimal in finite sample settings~\cite{rudelson2013hanson}.
Proofs are notoriously difficult, and simplifying them becomes an independent line of research~\cite{cohen2018simple,dasgupta2003elementary}.
\item \textbf{Hidden cost of gadgets.}
Several constructions are inherently statistically sub-optimal, trading the embedding variance for certain matrix patterns useful from the algorithmic perspective (such as sparsity~\cite{dasgupta2010sparse,li2006very}, circularity~\cite{cheng2014new}). 
\item \textbf{Lack of accurate impossibility results}. Related no-go results~\cite{kane2011almost,burr2018optimal}, concerning the relation of the confidence and  embedding dimension, hold asymptotically and are not accurate in finite-sample or fixed-accuracy regimes.
\end{itemize}
The goal of this work is to address these issues constructively. The challenge is:
\begin{framed}
\centering
Build random embeddings with statistically optimal accuracy. 
\end{framed}
The statistical accuracy is to be measured in terms of (a) variance or MSE of the distance distortion  (b) concentration properties (Berntein-type inequalities).

\subsection{Related Work}

Out of the large body of work following Johnson-Lindenstrauss random embeddings (e.g.~\cite{frankl1988johnson,indyk1998approximate,achlioptas2003database,ailon2006approximate,matouvsek2008variants,dasgupta2003elementary,dasgupta2010sparse,kane2014sparser,li2006very,freksen2018fully,cohen2016nearly,rojo2010improving}) 
only few papers prioritized statistically accurate analysis. As mentioned, several constructions are inherently sub-optimal trading variance for algorithmic properties~\cite{dasgupta2010sparse,li2006very,cheng2014new}, and many others doesn't offer transparent formulas. Best bounds are in ~\cite{achlioptas2003database,dasgupta2003elementary,rojo2010improving}.

The analysis can be reduced to the problem quadratic chaos estimation~\cite{krahmer2011new,braverman2010rademacher} with gain is in modularity and generality of sampling distributions, but at the cost of worse (or even non-transparent) numerical bounds due to dependencies.
Finally, the impossibility results for the Jonshon-Lindestrauss lemma~\cite{kane2011almost,burr2018optimal} do not establish accurate bounds for fixed data and embedding dimensions. 

\subsection{Contribution}

The novel contributions of this paper are:
\begin{itemize}
    \item \textbf{Sharp Variance and MSE Lower Bounds}, which gives the minimal possible variance or mean-squared error of the distortion in terms of matrix dimensions $m$ and $n$. This result is novel, and clarifies the fundamental limitation. Finding best-variance or best-mse estimation is of interest for any statistical problem~\cite{kiefer1952minimum}, but here it becomes even more relevant because of the connection to concentration bounds on the distortion.
    \item \textbf{Matching Construction} of random embeddings, achieving the best variance or mse. The construction builds on sampling from orthogonal matrices.
    \item \textbf{Exact Error Distribution} of the construction. Remarkably, we show that the error \emph{exactly} follows a beta distribution with explicit parameters; thus we avoid approximations as opposed to prior works. Furthermore, we show that this error obeys exponential (Bernstein-type) concentration guarantees, qualitatively similar to prior works, but with the optimal exponent.
    \item \textbf{Novel techniques} drawing on the \emph{singular eigenvalue decomposition}, which effectively replaces the use of Hanson-Wright lemmas, the bottlenecks in accuracy of prior analyses. Essentially, we are able to work with diagonal forms instead of a general quadratic chaos. We study the extreme properties of these diagonal expressions with  Schur-convexity and certain results on Dirichlet distribution, another novelty in the analysis of random embeddings. 
\end{itemize}

%The obvious drawback of statistical guarantees from prior works are \emph{input-agnostic}. They give same confidence bounds no matter if input data are sparse, dispersed or concentrated. According to empirical studies on random embeddings~\cite{venkatasubramanian2011johnson}, these theoretic bounds are not able to explain the observed, much better (!), behavior on real data. In this work we aim to bridge this gap,  by developing \emph{input-dependent} confidence guarantees. We solve the following problem

%The best tradeoff between the embedding dimension, distortion and confidence is achieved by the Rademacher sampler analyzed in~\cite{achlioptas2003database}. Recently it has been proved that that the constant in the bound cannot be improved under certain parameter regimes~\cite{burr2018optimal}, namely when the approximation error tends to $0$, the confidence tends to $1$, and the worst-case behavior of input data is considered. This result improved upon earlier, weaker, accuracy lower bounds~\cite{kane2011almost,jayram2013optimal,alon2003problems}. These lower bounds \emph{do not apply} to input-dependent confidence guarantees.

%None of this work has studied the more general case, where confidence is input-dependent.

\section{Results}

We consider fixed integers $1\leqslant n \leqslant m$, and a distribution $\mathcal{A}$ over matrices $n\times m$, refereed to as \emph{sampler}. For a sampler $\mathcal{A}$ the input data $x\in\mathbb{R}^m$ is projected as $A\cdot x$ where $A\sim\mathcal{A}$. The sampler distortion at point $x$ is
\begin{align}\label{eq:distortion}
E(x)\triangleq \|A x\|_2 / \|x\|_2 - 1,
\end{align}
which is the random variable depending on the sampled $A$. The sampling scheme is called unbiased when the distortion is unbiased
\begin{align}\label{eq:unbiased}
    \mathbf{E}_{A\sim\mathcal{A}}[E(x)] = 0.
\end{align}
If $\Pr_{A\sim\mathcal{A}}[|E(x)|>\epsilon]\leqslant \delta$, we say that the sampler $\mathcal{A}$
is $(\epsilon,\delta)$-DJL transform.

%This assumption is necessary to guarantee arbitrarily small distortions (the desired behavior). It is also present for technical reasons present in known analyses
%~\cite{frankl1988johnson,indyk1998approximate,achlioptas2003database,ailon2006approximate,matouvsek2008variants,dasgupta2003elementary,dasgupta2010sparse,kane2014sparser,li2006very,freksen2018fully,cohen2016nearly,rojo2010improving}. 

\subsection{Sharp Lower Bounds}

Below we determine what are \emph{best possible} variance and mean-squared error of DJL-transforms.
\begin{theorem}[Minimal Variance and MSE of DJL Transforms]\label{thm:min_max_variance}
Fix integers $1\leqslant n\leqslant m$, and let $x\in\mathbb{R}^m$ be sampled uniformly from the unit sphere.
Then it holds that:
\begin{align}
 \min_{\mathcal{A}:\text{unbiased}}\mathbf{Var}[E(x)] = \frac{2(m - n)}{n(m + 2)} ,
\end{align}
with the minimum over all unbiased samplers, and also:
\begin{align}
 \min_{\mathcal{A}}\mathbf{E}[E(x)^2] = \frac{2(m - n)}{m(n + 2)} 
\end{align}
with the minimum over all samplers. 
\end{theorem}
Note that both variance and mse are of order $O(1/n)$ when $1\ll n\ll m$.

\subsection{Matching Construction}

Best-variance and best-mse embeddings can be explicitly constructed as in \Cref{alg:best_variance_embedding}.
and \Cref{alg:best_mse_embedding}. Note that for sampling of orthogonal matrices one can use dedicated and efficient algorithms, as discussed in~\cite{anderson1987generation}.

\begin{algorithm}[H]
\SetKwInOut{Input}{input}
\SetKwInOut{Output}{output}
\DontPrintSemicolon
 \KwIn{feature dimension $m$, embedding dimension $n\leqslant m$.}
 \KwOut{best-variance random projection from $\mathbb{R}^m$ to $\mathbb{R}^n$. }
$V\sim^{uniform} \mathcal{O}(n) $ \tcp*{uniformly sampled orthogonal matrix $m\times m$}  
$\lambda\gets \sqrt{\frac{m}{n}}$\\
$\Lambda \gets \begin{bmatrix} \lambda\mathrm{diag}(\mathbf{1}_m) \\ \mathbf{0}_{n-m,m} \end{bmatrix}$ 
 \tcp*{diagonal matrix of shape $n\times m$} 
$U\sim^{any} \mathcal{O}(n)$ \tcp*{any orthogonal matrix $n\times n$}  
$A\gets U\Lambda V^T$\\
\Return $A$
\caption{Best-Variance JL Transform.}
\label{alg:best_variance_embedding}
\end{algorithm}
\begin{algorithm}[H]
\SetKwInOut{Input}{input}
\SetKwInOut{Output}{output}
\DontPrintSemicolon
 \KwIn{feature dimension $m$, embedding dimension $n\leqslant m$.}
 \KwOut{best-variance random projection from $\mathbb{R}^m$ to $\mathbb{R}^n$. }
$V\sim^{uniform} \mathcal{O}(n) $ \tcp*{uniformly sampled orthogonal matrix $m\times m$}  
$\lambda \gets \sqrt{\frac{(m+2)n}{2m+n^2}}$\\
$\Lambda \gets \begin{bmatrix} \lambda\mathrm{diag}(\mathbf{1}_m) \\ \mathbf{0}_{n-m,m} \end{bmatrix}$ 
 \tcp*{diagonal matrix of shape $n\times m$} 
$U\sim^{any} \mathcal{O}(n)$ \tcp*{any orthogonal matrix $n\times n$}  
$A\gets U\Lambda V^T$\\
\Return $A$
\caption{Best-MSE JL Transform.}
\label{alg:best_mse_embedding}
\end{algorithm}
Furthermore, we show that the error term is analytically tractable:
\begin{theorem}[Error Distribution of Optimal Embeddings]\label{thm:error}
Let $A$ be sampled as in \Cref{alg:best_variance_embedding}. Then the distortion is distributed as
\begin{align}
E(x) \sim \frac{m}{n}\cdot\mathsf{Beta}\left(\frac{n}{2},\frac{m-n}{2}\right)-1.
\end{align}
Let $A$ be sampled as in \Cref{alg:best_mse_embedding}. Then the distortion is distributed as
\begin{align}
E(x) \sim \frac{(m+2)n}{2m+n^2}\cdot\mathsf{Beta}\left(\frac{n}{2},\frac{m-n}{2}\right)-1.
\end{align}
\end{theorem}

For convenience we also show Bernstein-type concentration bounds, less precise than the Beta tail above, stated as the sub-gamma behavior. Note that the exponent for small $\epsilon$ becomes optimal (matching the variance).
\begin{corollary}[Optimal Bernstein-type Concentration]\label{cor:sub-gamma}
When $n=\Theta(m)$ we have the following sub-gamma behavior for \Cref{alg:best_variance_embedding}
\begin{align}
    E(x) \in \mathsf{SubGamma}(v^2,c),\quad v^2=\frac{2m}{m+2}\left(\frac{1}{n}-\frac{1}{m}\right), c = O(n^{1/2}),
\end{align}
which implies the tail bound for $\epsilon = o(n^{-1/2})$.
\begin{align}
    \Pr[\pm E(x) \geqslant \epsilon] \leqslant \mathrm{e}^{-\frac{\epsilon^2}{ v^2}(1+o(1)) }.
\end{align}
\end{corollary}

\subsection{Numerical Comparison}

\begin{figure}[h!]
\centering
\begin{subfigure}{.49\linewidth}
    \includegraphics[width=0.99\linewidth]{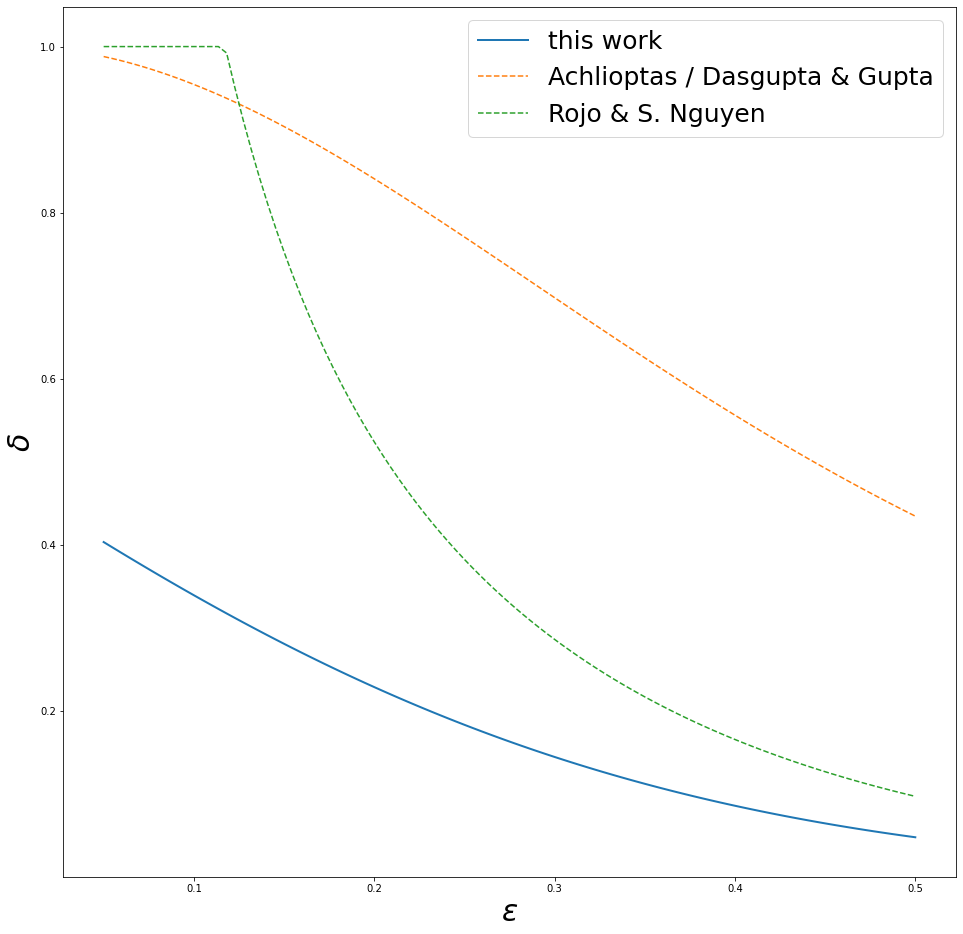}
    \caption{$m=10000,n=100$}
    \label{fig:1}
\end{subfigure}
\begin{subfigure}{.49\linewidth}
    \includegraphics[width=0.99\linewidth]{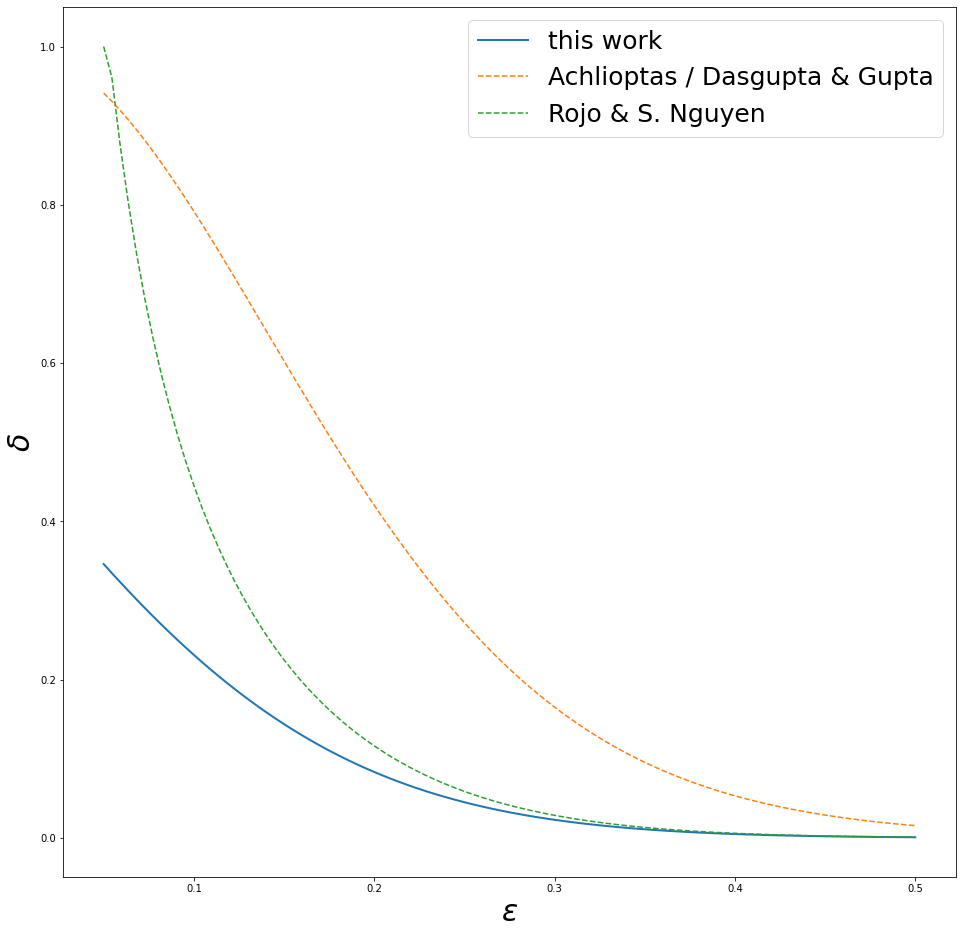}
    \caption{$m=100,n=20$}
    \label{fig:2}
\end{subfigure}
\caption{Comparison of our bounds with related work. Here $\delta = \Pr[E(x)>\epsilon]$ is the failure probability (smaller is better) and $\epsilon = E(x)$ is the distortion. \\
In \Cref{fig:1} the data dimension is reduced from $m=10000$ to $n=100$, 
in \Cref{fig:2} the data dimension is reduced from $m=100$ to $n=10$.}
\label{fig:comparison}
\end{figure}

In \Cref{fig:comparison} we present a detailed numerical comparison with bounds from related works.
For this comparison we select three best formulas, due to \cite{dasgupta2003elementary,achlioptas2003database,rojo2010improving}.

\subsection{Techniques}

Our calculations critically rely on the uniform distribution on the unit sphere. The strong added value of this work
is the following trick which \emph{effectively parametrizes} the uniform (Haar) measure on the sphere, using 
the elementary and explicit Dirichlet distribution. This result is novel:
prior works on impossibility results~\cite{kane2011almost,burr2018optimal} handled the unit sphere by a complicated three-term factorization, not a direct elementary distribution. As a stand-alone application of this technique, we simplify these results in a follow-up work.

\begin{lemma}[Parametrizing Sphere by Dirichlet Distribution]\label{lemma:sphere_dirichlet}
Let $X=(X_1,\ldots,X_m)$ be uniform on the unit sphere in $\mathbb{R}^m$. Then it holds that:
\begin{align}
(X_1^2,\ldots,X_m^2)\sim \mathsf{Dirichlet}\left(\frac{1}{2}\mathbf{1}_m\right).
\end{align}
\end{lemma}

The next powerful technique is parametrizing the distortion of a uniformly sampled unit vector;
we show that this distribution is a \emph{quadratic diagonal} expression, when conditioned on the singular values of the embedding matrix. This property will be critical to find close-form expressions for minimal variance/mse.

\begin{lemma}[Parametrizing Distortion by Singular Values]\label{lemma:distortion_singular}
Let $A =  U\Lambda V^T $ be the SVD decomposition of an $n\times m$ matrix $A$, where $n\leqslant m$. Let $X$ be uniform on  the unit sphere in $\mathbb{R}^{m}$. Then we have:
\begin{align}
    \|AX\|_2^2 \sim \sum_{k=1}^{n}\lambda_k^2 X_k^2
\end{align}
where $\lambda_1,\ldots,\lambda_n$ is the principal diagonal of $\Lambda$.
\end{lemma}
\begin{remark}
The expression depends only on the singular values and the input.
\end{remark}

Finally, we will need the notion of Schur-convexity, a powerful optimization technique. 
Armed with this tool we will be able to find best, under variance or mse, DJL transforms.
We say that $x\in\mathbb{R}^K$ is majorized by $y\in\mathbb{R}^K$, denoted by $x\prec y$ when $\sum_{i=1}^{k}x^{\downarrow}_i \leqslant \sum_{i=1}^{k}y^{\downarrow}_i$ for all $k=1\ldots K$.
A function $f$ in $K$ variables is called Schur-convex, abbreviated as S-convex, when 
$x\prec y$ implies
$f(x)\leqslant f(y)$.

\begin{lemma}[S-Convexity of Dirichlet Sums Variance]\label{lemma:variance_schur_convex}
Let $w,\ldots,w_n$ be non-negative weights and suppose that $(Z_1,\ldots,Z_m)\sim \mathsf{Dirichlet}\left(\lambda\mathbf{1}_m\right)$ for some $\lambda>0$, then for $n\leqslant m$ it holds that
\begin{align}
    F(w_1,\ldots,w_n)\triangleq\mathbf{Var}\left[\sum_{k=1}^{n}w_k Z_k\right]
\end{align}
is S-convex over the set of non-negative weights $(w_1,\ldots,w_n)$.
\end{lemma}

\section{Proofs}

\subsection{Proof of \Cref{thm:min_max_variance}} 

\subsubsection{Unbiased Embeddings}

Let $A$ be the sampled matrix (a random variable!) and $A=U\Lambda V^T$ be its singular value decomposition, 
that is $U$ and $V$ are orthogonal of shapes $n\times n$ and $m\times m$ respectively,
and $\Lambda$ is diagonal of shape $n\times m$ (they are, possibly correlated, random variables). Let
$\lambda_1,\ldots,\lambda_n$ is be the principal diagonal of $\Lambda$. By \Cref{lemma:distortion_singular}
\begin{align}
    \mathbf{Var}[\|AX\|_2^2 | \Lambda] = \mathbf{Var}\left[\sum_{k=1}^{n}\lambda_k^2 X_k^2\right].
\end{align}
Since $X$ is uniform on the sphere, we can use \Cref{lemma:sphere_dirichlet} to get the explicit form
\begin{align}
    \mathbf{Var}[\|AX\|_2^2 | \Lambda] = \mathbf{Var}\left[\sum_{k=1}^{n}\lambda_k^2 Z_k^2\right],\quad (Z_1,\ldots,Z_m)\sim\mathsf{Dirichlet}\left(\frac{1}{2}\mathbf{1}_m\right).
\end{align}
Using \Cref{lemma:variance_schur_convex} we find that the above expression is S-convex.
Denote $\bar{\lambda}^2 = \frac{1}{n}\sum_{k=1}^{n}\lambda^2_k$.
Clearly, we have the majorization
\begin{align}
\bar{\lambda}^2\mathbf{1}_n \prec (\lambda^2_1,\ldots,\lambda^2_n),
\end{align}
and therefore, the S-convexity implies
\begin{align}
   \mathbf{Var}[\|AX\|_2^2 | \Lambda] \geqslant  \mathbf{Var}\left[\sum_{k=1}^{n}\bar{\lambda}^2 Z_k^2\right]
   =\bar{\lambda}^4\mathbf{Var}\left[\sum_{k=1}^{n} Z_k^2\right].
\end{align}
By the properties of the Dirichlet distribution we have
\begin{align}
    \mathbf{Cov}[Z_i,Z_j] = \frac{2}{m(m+2)}\delta_{i,j}-\frac{2}{m^2(m+2)},
\end{align}
and, using this, we obtain
\begin{align}
   \mathbf{Var}[\|AX\|_2^2 | \Lambda] \geqslant \bar{\lambda}^4\left(\frac{2n}{m(m+2)}-\frac{2n^2}{m^2(m+2)}\right).
\end{align}
Recall that have also the constraint that $E(x) = \|Ax\|_2^2-1$ is unbiased for every $x$.
Then $\mathbf{E}[\|AX\|_2^2]=1$. Comparing this with \Cref{lemma:distortion_singular} we find that
\begin{align}
\begin{split}
 1 &= \mathbf{E}[\|AX\|_2^2] \\
& = \mathbf{E}_{\Lambda}\mathbf{E}\left[ \|AX\|_2^2 | \Lambda \right] \\
& = \mathbf{E}_{\Lambda}\left[ \sum_{k=1}^{n} \lambda_k^2 X_k^2 \right]. 
\end{split}
\end{align}
Using \Cref{lemma:sphere_dirichlet} we obtain
\begin{align}
    \mathbf{E}\left[\sum_{k=1}^{n} \lambda_k^2 X_k^2 \right] & =  \mathbf{E}\left[\sum_{k=1}^{n} \lambda_k^2 Z_k\right] \\
    & =  \frac{1}{m}\mathbf{E}\left[\sum_{k=1}^{n} \lambda_k^2 \right]
\end{align}
which, taking into account the definition of $\bar{\lambda}^2$ gives us
\begin{align}
    \mathbf{E}\bar{\lambda}^2 = \frac{n}{m}.
\end{align}
We use this in the variance lower bound and apply Jensen's inequality:
\begin{align}
\begin{split}
 \mathbf{E}_{\Lambda}\left[\mathbf{Var}[\|AX\|_2^2 | \Lambda]\right] & \geqslant \mathbf{E}[\bar{\lambda}^4]\left(\frac{2n}{m(m+2)}-\frac{2n^2}{m^2(m+2)}\right) \\
 & \geqslant \mathbf{E}[\bar{\lambda}^2]^2\left(\frac{2n}{m(m+2)}-\frac{2n^2}{m^2(m+2)}\right)\\
 & = n^2 m^{-2}\left(\frac{2n}{m(m+2)}-\frac{2n^2}{m^2(m+2)}\right) \\
 & = \frac{2m}{m+2}\left(\frac{1}{n}-\frac{1}{m}\right).
\end{split}
\end{align}
Finally, by the total variance law, and the fact that variance of any random variable is non-negative:
\begin{align}
\begin{split}
\mathbf{Var}[\|AX\|_2^2] & = \mathbf{E}_{\Lambda}\left[\mathbf{Var}[\|AX\|_2^2 | \Lambda]\right] + \mathbf{Var}[\mathbf{E}[
 \|AX\|_2|\Lambda]] \\
 & \geqslant \mathbf{E}_{\Lambda}\left[\mathbf{Var}[\|AX\|_2^2 | \Lambda]\right]
\end{split}
\end{align}
which, taken into account the previous step, finishes the proof because $E(X) = \|AX\|_2^2-1$.
It remains to comment on the optimality: a) the Schur and Jensen inequalities are sharp when $\lambda_k$ are equal,
and b) skipping the term in the total-variance is sharp when $\mathbf{E}[
 \|AX\|_2|\Lambda]$ does not depend on $\Lambda$, for instance when $\lambda_k$ is deterministic.

\subsubsection{Biased Embeddings}

We follow the derivation as in the unbiased case and use same notation. With no changes we arrive at:
\begin{align}
   \mathbf{Var}[\|AX\|_2^2 | \Lambda] \geqslant \bar{\lambda}^4\left(\frac{2n}{m(m+2)}-\frac{2n^2}{m^2(m+2)}\right).
\end{align}
By the total variance law and Jensen's inequality:
\begin{align}
\begin{split}
\mathbf{Var}[\|AX\|_2^2] & \geqslant \mathbf{E} \left[\mathbf{Var}[\|AX\|_2^2 | \Lambda]\right] \\
& \geqslant \mathbf{E}[\bar{\lambda}^4]\left(\frac{2n}{m(m+2)}-\frac{2n^2}{m^2(m+2)}\right) \\
& \geqslant \mathbf{E}[\bar{\lambda}^2]^2\left(\frac{2n}{m(m+2)}-\frac{2n^2}{m^2(m+2)}\right).
\end{split}
\end{align}
This time we have to consider the contribution from the bias term. By \Cref{lemma:distortion_singular}
\begin{align}
    \mathbf{E}[\|A X\|_2^2-1 | \Lambda] = \sum_{k=1}^{n} \lambda_k^2 X_k^2,
\end{align}
and by \Cref{lemma:sphere_dirichlet} we get
\begin{align}
    \mathbf{E}[\|A X\|_2^2-1 | \Lambda] = \sum_{k=1}^{n} \lambda_k^2 Z_k, \quad (Z_1,\ldots,Z_m)\sim \mathsf{Dirichlet}\left(\frac{1}{2}\mathbf{1}_m\right).
\end{align}
By averaging over $\Lambda$ and using the fact that $\mathbf{E}Z_k = \frac{1}{m}$ we obtain:
\begin{align}
\begin{split}
    \mathbf{E}[\|A X\|_2^2-1 ]&= \mathbf{E}_{\Lambda}[\mathbf{E}[\|A X\|_2^2-1 | \Lambda] ]\\
    & =\mathbf{E}_{\Lambda}\left[\mathbf{E}\left[\sum_{k=1}^{n} \lambda_k^2 Z_k-1| \Lambda\right]\right]-1 \\
    & = \frac{1}{m}\mathbf{E}\left[\sum_{k=1}^{n}\lambda_k^2\right]-1 \\
    & = \frac{n}{m}\mathbf{E}[\bar{\lambda}^2]-1.
\end{split}   
\end{align}
By the bias-variance decomposition we have that
\begin{align}
    \mathbf{E}[(\|AX\|_2^2-1)^2]=\mathbf{Var}[\|AX\|_2^2] +  \left(\mathbf{E}[\|A X\|_2^2-1 ]\right)^2.
\end{align}
We combine this with the bound on the variance, and the bias expression:
\begin{align}
    \mathbf{E}[(\|AX\|_2^2-1)^2]\geqslant \mathbf{E}[\bar{\lambda}^2]^2\left(\frac{2n}{m(m+2)}-\frac{2n^2}{m^2(m+2)}\right) + \left(\frac{n}{m}\mathbf{E}[\bar{\lambda}^2]-1.\right)^2.
\end{align}
Consider the function $g(u)=u^2\left(\frac{2n}{m(m+2)}-\frac{2n^2}{m^2(m+2)}\right) + \left(\frac{n}{m}u-1\right)^2$, so that the above bound equals $g(\mathbf{E}[\bar{\lambda}^2])$.
The minimum is achieved for $u = \frac{(m+2)n}{2m+n^2}$, as we find with the help of \texttt{Sympy} package~\cite{sympy}. The corresponding minimal value is $g(u) = \frac{2(m - n)}{m\cdot(n + 2)}$ (see below for the code). The bound is sharp when $\lambda_k = \frac{(m+2)n}{2m+n^2}$.
\begin{lstlisting}[language=Python,frame=single]
from sympy import Symbol, diff
from sympy.solvers import solve

m = Symbol('m')
n = Symbol('n')
x = Symbol('x')

var = x**2*(2*n/(m*(m+2))-2*n**2/(m**2*(m+2)))
bias = n/m*x-1
mse = var+bias**2

x_best = solve(diff(mse,x),x)[0]

mse_opt = mse.subs({x:x_best}).simplify()
var_unb = var.subs({x:m/n}).simplify()

print(var_unb)
print(mse_opt)
\end{lstlisting}

\subsection{Proof of \Cref{thm:error}}

Consider $A$ sampled according to \Cref{alg:best_variance_embedding} or \Cref{alg:best_mse_embedding}.
We have $A = U\Lambda V^T$
where $\Lambda$ is rectangular-diagonal with diagonal entries 
$\lambda_k=\lambda$, where $\lambda = \sqrt{\frac{m}{n}}$ for the best-variance case
respectively $\lambda = \sqrt{\frac{(m+2)n}{2m+n^2}}$ for the best-mse case, $U$ is orthogonal matrix
of shape $n\times n$ and $V$ is an independent, uniformly sampled  orthogonal matrix of shape $n\times n$. 
For every $x$ we have $\|Ax\| _2= \|U\Lambda V^Tx\|_2 = \|\Lambda V^Tx\|_2$. 
Denoting $w=V^T x$ we can write
\begin{align}
E(x)=\lambda^2\sum_{k=1}^{n}w_i^2-1.
\end{align}
When $x\in S^{m-1}$ and $V$ is uniformly orthogonal, the vector $w_i = V^Tx$ is uniformly distributed on $S^{m-1}$. Denoting $z_i = w_i^2$ we therefore obtain
\begin{align}\label{eq:explicit_error}
E(x)= \lambda^2\sum_{i=1}^{n}z_i - 1,\quad (z_1,\ldots,z_m)\sim \mathsf{Dirichlet}\left(\frac{1}{2}\mathbf{1}_m\right).
\end{align}
By the marginal properties of Dirichlet distribution~\cite{albert2012dirichlet} we conclude that
\begin{align}
\frac{E(x)+1}{\lambda^2} \sim \mathsf{Beta}\left(\frac{n}{2},\frac{m-n}{2}\right),
\end{align}
as required.

\subsection{Proof of \Cref{cor:sub-gamma}}

Denote $Y = \mathsf{Beta}\left(\frac{n}{2},\frac{m-n}{2}\right)$. By the sub-gaussian property of the beta distribution~\cite{zhang2020non} we have (up to an absolute constant)
\begin{align}
    \|Y-\mathbf{E}Y\|_d \leqslant O(\sqrt{d/m}),
\end{align}
which in turn implies
\begin{align}
\|E(x)\|_d \leqslant O(\sqrt{d}\cdot \sqrt{m}/n).
\end{align}
Using this, and the fact that $\mathbf{E}(x)=0$, we can estimate the MGF
\begin{align}
\begin{aligned}
    \mathbf{E}\exp(t E(x)) &= 1 + \frac{\mathbf{Var}[E(x)]t^2}{2} + \sum_{d\geqslant 3}\frac{O(\sqrt{d}\cdot \sqrt{m}/n)^d t^d}{d!} \\
    & \leqslant 1 + \frac{\mathbf{Var}[E(x)]t^2}{2} + \frac{t^3m^{3/2}}{n^3}\cdot \exp(O(m t^2/n^2)).
\end{aligned}
\end{align}
Now, assuming that $0.01\leqslant m\leqslant 0.99n$, we have $\mathbf{Var}[E(x)] = \Theta(1/n)$, and 
\begin{align}
\mathbf{E}\exp(t E(x)) &\leqslant 1 + \frac{\mathbf{Var}[E(x)]t^2}{2}\cdot\left(1+O(t/n^{1/2})\right),\quad\text{if } |t| = O(n^{1/2}).
\end{align}
It follows that for some constant $c$ we have
\begin{align}
\mathbf{E}\exp(t E(x)) \leqslant 1 + \frac{\mathbf{Var}[E(x)]t^2}{2}\cdot(1+c|t|n^{1/2}) ,\quad c|t|n^{1/2}<1,
\end{align}
and thus
\begin{align}
\mathbf{E}\exp(t E(x)) \leqslant 1 + \frac{\mathbf{Var}[E(x)]t^2}{2\cdot(1-c|t|n^{1/2})} ,\quad c|t|n^{1/2}<1.
\end{align}
Using the logarithmic inequality~\cite{love198064} we finally get
\begin{align}
\log\mathbf{E}\exp(t E(x)) \leqslant \frac{\mathbf{Var}[E(x)]t^2}{2}\cdot(1-c|t|n^{1/2}), \quad c|t|n^{1/2}<1,
\end{align}
so that we conclude $E(x)\in \Gamma(\mathbf{Var}[E(x)],O(n^{1/2}))$, as required. The tail bound follows by the sub-gamma properties.

\bibliographystyle{splncs04}
\bibliography{citations}

\appendix

\section{Proofs}

\subsection{Proof of \Cref{lemma:sphere_dirichlet}}

Recall that the uniform vector on the sphere $(Z_1,\ldots,Z_m)$ can be sampled as 
$Z_i = N_i / \sqrt{\sum_{j=1}^{K} N_j}$ where $N_j$ are iid zero-mean normal. Take $N_j\sim\mathsf{Norm}(0,1/2)$
then we have $\Gamma_i=N_i^2 \sim\mathsf{Gamma}\left(\frac{1}{2},1\right)$. The result follows now, by the fact that
the  components of the Dirichlet distribution with parameters $(\alpha_1,\ldots,\alpha_m)$ can be parametrized as
$(\Gamma_i)_i/\sum_{j=1}^{m}\Gamma_j$ where $\Gamma_j\sim \mathsf{Gamma}(\alpha_i,1)$.

\subsection{Proof of \Cref{lemma:distortion_singular}}
\begin{proof}
Consider the singular eigenvalue decomposition $A = U\Lambda V^T$
where $\Lambda$ is a diagonal and non-negative matrix of shape $n\times m$, and $U,V$ are orthogonal matrices of shapes $m\times m$ and $n\times n$ respectively.
Let $w = V^T x$, it is of unit length and $\lambda_i$ be the diagonal entries of $\Lambda$. Then
\begin{align}
 \|Ax\|_2^2 = \|U\Lambda w\|_2^2 = \|\Lambda w\|_2^2 = \sum_{k=1}^{n} \lambda_k^2 w_k^2.
\end{align}
Let $X$ be uniform on the unit sphere $S^{m-1}$ and denote $W = V^T X$. Since $V^T$ is orthogonal, $W$ is also uniform on $S^{m-1}$, as required.
%Then, by \Cref{lemma:sphere_dirichlet}, we know that the vector $(W_i^2)_i$ is distributed as $\mathsf{Dirichlet}\left(\frac{1}{2}\cdot \mathbf{1}_m\right)$. Thus
%\begin{align}
%    \|AX\|_2^2-1 \sim \sum_{k=1}^{n} Z_k,\quad (Z_1,\ldots,Z_n)\sim \mathsf{Dirichlet}\left(\frac{1}{2}\mathbf{1}_m\right),
%\end{align}
%as required.
\end{proof}

\subsection{Proof of \Cref{lemma:variance_schur_convex}}

By the variance formula and the variance properties of the Dirichlet distribution
\begin{align}
\begin{split}
    \mathbf{Var}[\sum_{k=1}^{n}w_k Z_k] &= \sum_{1\leqslant i,j\leqslant n}w_i w_j\mathbf{Cov}[Z_i,Z_j]\\
    & = \sum_{i=1}^{n}\frac{w_i^2}{m(1+m\lambda)} - \sum_{1\leqslant i,j\leqslant n}  \frac{w_i w_j}{m^2(1+m\lambda)}.
\end{split}
\end{align}
For readability, we introduce the matrix 
\begin{align}
Q = \frac{1}{m(1+m\lambda)}\cdot \mathrm{diag}(\mathbf{1}_n) - \frac{1}{m^2(1+m\lambda)}\cdot\mathbf{1}_{n,n}.
\end{align}
Then, denoting $w = (w_1,\ldots,w_n)$, we can write:
\begin{align}
        \mathbf{Var}\left[\sum_{k=1}^{n}w_k Z_k\right] = w^T Q w.
\end{align}
We now prove that $Q$ is positive semi-definite. To see that,
note that $Q$ is proportional, by a positive scalar, to the difference $\mathrm{diag}(\mathbf{1}_m)-\frac{1}{n}\mathbf{1}_{m,m}$.
Clearly, $Q$ is symmetric and has real eigenvalues. The eigenvalues of the first matrix are $1$ repeated $m$ times, and the biggest eigenvalue
of the second matrix equals $\frac{m}{n}$. By the Weyl result on perturbation of symmetric (more generally: complex hermitian) matrices~\cite{weyl1912asymptotische}, the eigenvalues of $Q$ are at least $1-\frac{m}{n}\geqslant 0$, 
here we use the assumption $n\leqslant m$. Thus, we have proved that $Q$ is positive semi-definite 
(note: we could also use a result on rank-one perturbations~\cite{golub1973some}).

If $Q$ is positive semi-definite, its quadratic form is a convex function. Since it is symmetric and the unit simplex
is symmetric, we conclude that it is also S-convex~\cite{law2007effective}.

\end{document}